\documentclass[times,final]{article}
    \makeatletter
    \def\ps@pprintTitle{%
    \let\@oddhead\@empty
    \let\@evenhead\@empty
    \def\@oddfoot{}%
    \let\@evenfoot\@oddfoot}
    \makeatother
    \usepackage{lineno}
    \usepackage{mathrsfs}
    \usepackage[breaklinks=true,pdftex]{hyperref}
    \modulolinenumbers[1]
    \usepackage{amsmath,bm}
    \usepackage{svg}
    \usepackage{comment}
    \usepackage{multirow}
    \usepackage[a4paper,
            bindingoffset=0.1in,
            left=1.5cm,
            right=1.5cm,
            top=1.5cm,
            bottom=1.5cm]{geometry}
    \usepackage{ulem}
    \usepackage{booktabs}
    \usepackage{dirtytalk}
    \usepackage{lineno}
    \usepackage{caption}
    \usepackage{amssymb}
    \usepackage{amsmath}
    \usepackage{savesym}
        \usepackage{algorithmic}
    \usepackage{algorithm}
    \savesymbol{AND}
    \usepackage{adjustbox}

    \usepackage{rotating}
    \usepackage{graphicx}
    \usepackage{wrapfig}
    \usepackage{amssymb}
    \usepackage{soul,color}
    \usepackage{subcaption}
    \usepackage{lineno}
\usepackage{xcolor}
\usepackage{url}
    \soulregister\cite7
    \soulregister\ref7
    \soulregister\pageref7

    \usepackage{todonotes}

    \DeclareRobustCommand{\uvec}[1]{{%
    \ifcsname uvec#1\endcsname
    \csname uvec#1\endcsname
    \else
    \bm{\hat{\mathbf{#1}}}%
    \fi}}
\mathchardef\breakingcomma\mathcode`\,
{\catcode`,=\active
  \gdef,{\breakingcomma\discretionary{}{}{}}
}

\begin{document}

\begin{center}
{\LARGE \textbf{How does agency impact human-AI collaborative design space exploration? A case study on ship design with deep generative models}}\\
\vspace{0.5cm}
{\small Shahroz Khan$^{1,2,*}$\let\thefootnote\relax\footnote{$^*$Corresponding author. E-mail address: shahroz.khan@strath.ac.uk; shahroz.khan@berkeley.edu (S. Khan)}},
{\small Panagiotis Kaklis$^1$},
{\small Kosa Goucher-Lambert$^2$}
\\\vspace{0.2cm}
{\small $^1$Department of Naval Architecture, Ocean and Marine Engineering, University of Strathclyde, Glasgow, United Kingdom}\\
{\small $^2$Department of Mechanical Engineering, University of California, Berkeley, United States}
\end{center}

\section*{\centering Abstract}
Typical parametric approaches restrict the exploration of diverse designs by generating variations based on a baseline design. In contrast, generative models provide a solution by leveraging existing designs to create compact yet diverse generative design spaces (GDSs). However, the effectiveness of current exploration methods in complex GDSs, especially in ship hull design, remains unclear. To that end, we first construct a GDS using a generative adversarial network, trained on 52,591 designs of various ship types. Next, we constructed three modes of exploration, random (REM), semi-automated (SAEM) and automated (AEM), with varying levels of user involvement to explore GDS for novel and optimised designs. In REM, users manually explore the GDS based on intuition. In SAEM,  both the users and optimiser drive the exploration. The optimiser focuses on exploring a diverse set of optimised designs, while the user directs the exploration towards their design preference. AEM uses an optimiser to search for the global optimum based on design performance. Our results revealed that REM generates the most diverse designs, followed by SAEM and AEM. However, the SAEM and AEM produce better-performing designs. Specifically, SAEM is the most effective in exploring designs with a high trade-off between novelty and performance. In conclusion, our study highlights the need for innovative exploration approaches to fully harness the potential of GDS in design optimisation.
\vspace{0.2cm}\\
\textit{Keywords:} Design Exploration, Generative AI, Human-AI Collaboration, Shape Optimisation, Ship Design
\section{Introduction}
The ability of a design space to create a rich and valid set of design alternatives is a crucial component of shape optimisation pipelines, as it determines the quality and innovativeness of the solutions produced. Typically, design spaces result from the parametric modellers, which are pre-coded to parameterise the key features of a baseline design \cite{intra_r56}. However, these modellers are built to produce solutions within the proximity of the baseline design, and thus, have several limitations. One such drawback is the limited ability of the resulting design spaces to support rich design exploration, leading to a lack of novel design solutions \cite{chen2021padgan}.

Moreover, machine learning approaches have proven effective in bypassing the need for computational solvers by providing low-fidelity performance estimators trained offline with data from high-fidelity solvers or physical experimentation. However, until recently, the ability of these models to generate innovative solutions was limited. This limitation arose from the fact that they were only built to predict the performance criteria of designs coming from very narrow design spaces. Generative models \cite{regenwetter2022deep} such as generative adversarial networks (GANs), variational auto-encoders (VAEs), diffusion models, and transformers have changed this by providing rich design spaces that allow for the creation of innovative shapes in addition to performance prediction.

In engineering design tasks, generative models are gaining attention for creating vast generative design spaces (GDSs). These models learn a set of latent features from the given training dataset of existing designs, which are used as design parameters to form GDSs. GDSs are not only low dimensional to expedite shape optimisation but, if properly trained, can also produce novel and valid design alternatives beyond the spectrum of the training dataset. Additionally, efforts are underway to enhance the quality of GDSs to make them physics-informed \cite{yang2020physics} and user-centred \cite{chaudhari2023evaluating}. Physics-informed GDSs can leverage physical laws to ensure that generated designs satisfy certain performance criteria, while user-centred GDSs can incorporate user preferences and constraints to generate designs that are more aligned with the user's needs. 

Although GDSs have the potential to offer unprecedented design possibilities, their usability in real design scenarios is not yet fully understood. It is crucial to determine how GDSs can be best utilised without overwhelming designers while expediting the design process. For example, it is essential to study whether existing design exploration techniques, primarily designed to explore narrow design spaces generated by procedural parametric modellers  \cite{khan2017novel,intra_r56}, can be effectively applied to explore vast design spaces offered by generative models. Moreover, it is important to understand whether designers are willing to adopt GDSs in their design activities and, if so, what innovative design approaches and scenarios they can use to take full advantage of these diverse spaces. 

To achieve this understanding, the present work investigates the most efficient ways of design exploration of GDSs. To this end, we first construct a GDS for complex engineering design problems, such as ship hull design, where parametric design plays a vital role. We create the GDS for hull design by training a custom GAN model, ShipHullGAN \cite{ShipHullGAN_2023}, on a large dataset of various ship types, including tankers, container ships, bulk carriers, tugboats, and crew supply vessels.

We then develop three design exploration modes with varying degrees of autonomy or designer involvement: random, semi-automated, and automated. The random exploration mode (REM) is a typical preliminary design phase, where designers independently explore the design space based on their intuition and expertise while considering performance. In the semi-automated exploration mode (SAEM), both the designer and optimiser collaborate to guide design exploration towards user-centred and optimised areas of GDS. Finally, the automated exploration mode (AEM) is the standard optimisation scenario where the optimiser is the primary driver and design space exploration occurs while taking performance into account.

With the above modes of exploration, we aim to understand \textit{how designers/naval architects perceive different modes of design exploration in the quest of generating novel design solutions from GDS}. With the above research question in mind, during the study, we aim to mainly analyse the following: 

\begin{enumerate}
    \item To what extent is each exploration mode effective in achieving diverse, novel and better-performing designs?
    \item Which factor is the key consideration for each exploration mode: form or performance?
\end{enumerate}

\section{Background on ship design and optimisation}
Ship design is a complex and bespoke engineering process \cite{charisi2022early}, which differs significantly from other design fields. Unlike other industries, there is no opportunity for full-scale testing, which means that designers have to rely heavily on digital design tools to create the most efficient and safe vessels possible \cite{nowacki2010five}. In today's highly competitive world market, ships must be designed to meet high standards while also being delivered quickly. This requires a high degree of optimisation and customisation, as designers must balance numerous factors such as fuel efficiency, speed, safety, and cargo capacity \cite{gaspar2012addressing,ebrahimi2021influence}. The ultimate objective of ship design is to achieve the best performance for a given set of design criteria, which includes the vessel's intended use, the environmental conditions it will operate in, and the regulatory requirements that it must meet. Achieving these objectives requires a multidisciplinary approach that combines expertise in naval architecture, marine engineering, materials science, and other fields \cite{papanikolaou2010holistic}.

To expedite the design process, naval architects use extensively off-the-shelf parametric modelling tools. These tools are characterised by conservatism, for they are built to generate shapes lying in the neighbourhood of a successful baseline/parent shape \cite{khan2017customer}. Some relevant examples of such tools are presented in \cite{GinnisEtAl2011,khan2022modiyacht,intra_r56,khan2019genyacht,khan2017novel}. Next, these modellers are coupled with optimisers for improving the baseline shape against performance criteria (e.g., ship wave resistance, seakeeping, structural strength, etc.), which involve time-consuming simulations, e.g., computational fluid dynamics (CFD). At the end of the process, the new design is likely a local optimum whose shape is a minor variation of the existing one. While these approaches have proven effective for well-established ship types, there may be a need for more radical design ideas in certain situations. This can occur in situations where there are specific requirements that necessitate a more extensive exploration of the design space. Additionally, it may arise when there is a need to revolutionize and redesign existing ship types due to significant regulatory changes, such as the IMO 2020 emission reduction mandate, or the emergence of new disrupting technologies in the context of Industry 4.0 \cite{DKaklisEtAl2023SOME,citaristi2022united,joung2020imo}. Such a strategy will benefit novel design tasks, e.g., special-purpose vessels, but it can also offer a competitive advantage for traditional players in the industry.

Conclusively, the coexistence of conservative parametric modellers with high-cost simulations and a large number of design parameters needed for shape optimisation of complex shapes leads to a non-efficient design approach. Such an approach can suffer from the curse of high dimensionality and a limited capability to explore design spaces efficiently for delivering variant, innovative, user-centred and truly optimal designs \cite{khan2022shape}. 

Therefore, the ship design necessitates design approaches those bypass the dependence on the parent design and use more rational methods to create rich design spaces, i.e., design spaces resulting from the generative models, with the ability to formulate both conventional and non-conventional hull forms \cite{ShipHullGAN_2023}. 

\section{Research methodology}
For this work, a study involving human subjects has been developed to quantitatively analyse the efficiency of three exploration modes for exploring GDS constructed using a custom GAN for the ship hull design. Firstly, we discuss the construction of the GDS and how it can be used in preliminary optimisation while being connected to a surrogate model to predict design performance. We then provide a detailed discussion on the different modes of exploration used to analyse the performance of GDS and how they differ in terms of optimisation and user involvement.

\subsection{Creation of generative design space (GDS)}
There have been substantial efforts in computer-aided ship design for building robust parametric tools, but they can only handle a specific hull type \cite{khan2017novel,intra_r56}. Despite their efficiency in creating valid and smooth ship-hull geometries, they cannot be readily used to generate instances of ship types that deviate significantly from their target ship types. Therefore, in this work, we utilised, ShipHullGAN \cite{ShipHullGAN_2023}, a generic parametric modeller built using deep convolutional GANs. The training of ShipHullGAN is performed using a large and diverse dataset of existing hull geometries. We first extensively explored the literature on hull form optimisation and machine learning to identify various hull types. Ultimately, we selected 17 different parent hulls, including KCS \footnote{\url{http://www.simman2008.dk/KCS/kcs_geometry.htm}}, KVLCC2 \footnote{\url{http://www.simman2008.dk/kvlcc/kvlcc2/kvlcc2_geometry.html}}, VLCC, JBC \footnote{\url{https://www.t2015.nmri.go.jp/jbc.html}}, DTC, DTMB \footnote{\url{http://www.simman2008.dk/5415/combatant.html}}, and others from the FORMDATA series. We then created 3,000 synthetic variations of these hulls using the parametric approach described in \cite{intra_r56}. The length, beam, and width of these designs were kept constant, while non-dimensional parameters between 0 and 1 were used to create shape variations. For the FORMDATA series, 5000 design variations were created systematically with respect to the characteristic parameters like midship section-area coefficient $c_M$ and the block coefficients $c_{BA}$ and $c_{BF}$ of the aft and fore parts of the ship, respectively. This synthetic and systematic design creation resulted in 56,000 designs. Subsequently, in order to establish a reliable training dataset, designs undergo validation using a blend of geometry- and physics-oriented quality filters to assess the viability of each design. Geometry-based filters are employed to ascertain geometric validity, ensuring the absence of self-intersecting surfaces in all designs. Conversely, physics-based filters are utilised to verify that the performance of each design can be accurately predicted by solvers without any potential collapse. Following this rigorous design validation process, a total of 52,591 design variations, which have been both geometrically and physically validated, are obtained for training the ShipHullGAN model.

The design dataset to the ShipHullGAN is inputted in the form of a shape-signature vector (SSV), which consists of
a shape modification function and geometric moments. SSV acts as a unique descriptor of each dataset design instance \cite{khan2022shape,khan2022geometric}. The inclusion of geometric moments enables the extraction of meaningful features that are not only geometry-driven but also physics-informed. Using geometric moments along with the shape increases the chances of creating a large number of geometrically valid shapes, as adding moments gives a rich set of information about the geometry. More importantly, a strong correlation between ship physics and geometric moments also induces the notion of physics in the extracted latent features. Thus, the resulting features have not only the ability to form a compact but also a physics-informed design, ensuring high-quality valid designs.  

The ShipHullGAN uses deep convolutional neural networks for both generator ($G$) and discriminator ($D$) components to capture sparsity in the training dataset, along with a space-filling term in the loss function to enhance diversity. $D$ consists of 6 convolutional layers and a dropout layer, with a sigmoid activation function in the last convolutional layer to determine if the design is real or fake. $G$ is the transpose of $D$ and has 5 transposed convolutional layers, with an input layer that takes randomly sampled design, $\mathbf{x}$ and reshapes it. Both $G$ and $D$ use batch normalisation and ReLU activation functions. Training is performed using the Adam gradient descent algorithm with specific settings and performed on a computer with a dual 24-core 2.7GHz Intel Xeon 6 Gold 6226 CPU, NVIDIA Quadro RTX 6000 GPU, and 128GB of memory.

\begin{figure}[htb!]
    \centering
    \includegraphics[width=0.4\textwidth]{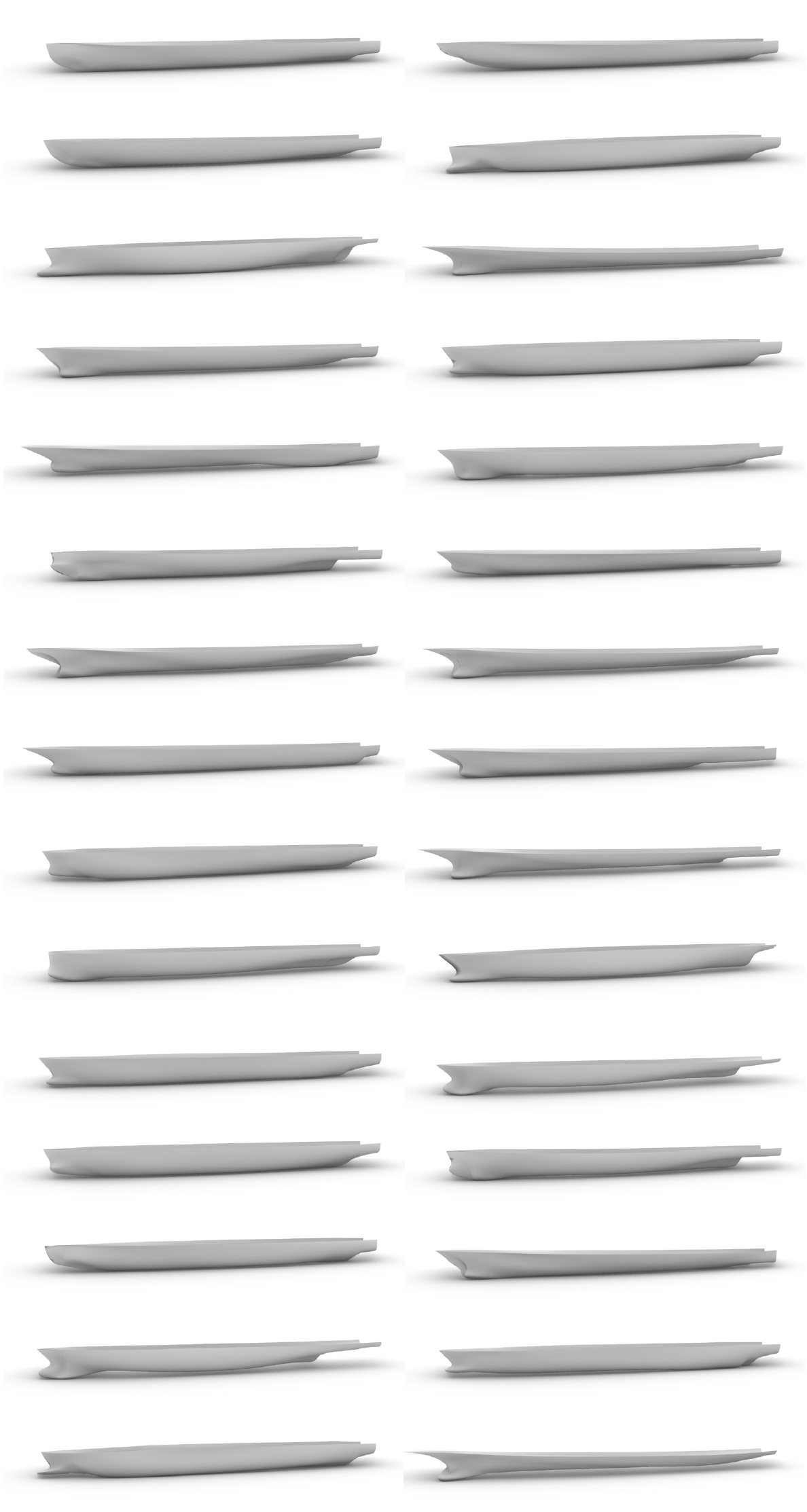}
    \caption{Design variations created with the proposed parametric modeller. These design variations can be visualised at \url{https://youtu.be/avlq0FxZP-s} and \url{https://youtu.be/ZIfmAs5-qFw}}
    \label{f1}
    \end{figure}

Once the training is completed, the generator component of the ShipHullGAN model is used as a generic parametric modeller. This provides a rich 20-dimensional GDS, which facilitates users in exploring design variations for a wide range of ship hulls. The resulting design variations include both traditional and unconventional forms, as shown in Figure \ref{f1}. Interested readers should refer to \cite{ShipHullGAN_2023} for details on the training of ShipHullGAN and the construction of 20-dimensional GDS. 

\subsection{Optimisation}
For the three modes of exploration, a simple optimisation problem is formulated. The problem aims to explore the 20-dimensional GDS resulting from the ShipHullGAN parametric modeller to create a container ship with a load-carrying capacity of 3600 TEU (Twenty-foot equivalent unit) while minimising its wave-making resistance/drag ($C_w$). This optimisation problem can be written in the following setting:

    \begin{equation}\label{shipganOptEq1}
    \begin{aligned}
    \textrm{Find } \mathbf{x^*}\in\mathbb{R}^{20} \quad & \textrm{such that} \\
    C_w(\mathbf{x^*}) = &  \min_{\mathbf{x} \in \mathcal{X}} C_w(\mathbf{x}) \\
    \textrm{subject to:}\quad & \textrm{given cargo capacity (3600 TEU)};\\
    \quad & 51120.5m^3 \leq \text{Volume of displacement} \leq 56501.6  m^3;\\
    \quad & 220.9m \leq \text{Length at waterline} \leq 244.2m;\\
    \quad & 30.6m \leq \text{Beam at waterline} \leq 33.8 m;\\
    \quad & 10.3m \leq \text{Draft} \leq 11.3m.
     \end{aligned}
    \end{equation}

\noindent The design constraints in the above equation are set to obtain physically plausible variations of the hull designs. The physical criterion, $C_w$, is part of the overall resistance affecting the movement of objects on or near the free surface of oceans, lakes and rivers. It reflects the energy spend on creating the free-surface waves following the moving body \cite{bertram2011practical}. Although the overall resistance of the ship is composed of different components, $C_w$ is a vital component and especially prominent for relatively full hull forms travelling at high speeds. It is noteworthy that $C_w$ is highly sensitive to local features of the hull so that a significant reduction can be achieved without affecting the overall cargo capacity. $C_w$ is affected by the distribution of the hull's shape, and minimising it at the preliminary design stage is crucial, but its evaluation can be highly computationally demanding. 

\subsection{Performance evaluation}
To expedite the optimisation process and reduce user fatigue resulting from long simulation run times, we developed a surrogate model that predicts $C_w$ values for designs using Gaussian Process regression (GPR) \cite{schulz2018tutorial}. GPR is a non-parametric Bayesian approach that has been used in various design applications. It maps the globally-coupled, non-linear relationship between inputs and outputs sampled from a theoretically infinite-dimensional normal distribution and any finite number of input-space samples that follow a corresponding joint (multivariate) Gaussian distribution. The main advantages of GPR over other modelling techniques are that it can: (1) map the input-output relationship with small data size, (2) handle noise in the data easily, thus avoiding over-fitting, and (3) optimise hyperparameters from training data to increase the fit accuracy.

To develop a reliable GPR model, we sampled 10,000 designs using the dynamic propagation sampling technique \cite{khan2021regional}, which ensures that designs are evenly distributed in the design space covering all the design possibilities a given design has to offer. For evaluating $C_w$ values of the designs in the training dataset, we performed hydrodynamic simulations using a software package based on linear potential flow theory using Dawson (double-model) linearisation, with details of the employed formulation, the numerical implementation, and its validation appearing in \cite{bassanini1994wave}. As a result of using simple Rankine sources, the computational domain consists of a part of the undisturbed free surface, extending 1$Lpp$ upstream,  3$Lpp$ downstream, and 1.5$Lpp$ sideways, with $Lpp$ denoting the length between perpendiculars for the assessed ship hull. A total of $[20 \times70]$ grid points are used for the undisturbed free surface, whereas  $[50\times180]$ grid points are used for the hull discretisation with the simulation being performed at a Froude number $F_r$ equal to $F_r=U/\sqrt{gL}=0.28$, where $g$ is the acceleration due to gravity, and $L$ is the ship's length. Readers can refer to \cite{khan2021physics} for details on the construction of the surrogate model with GPR. 

\subsection{Experiment procedures}
The study is composed of three generative design exploration modes, random, semi-automated and automated design exploration, with varying levels of user involvement while providing them with a different level of autonomy. In the following section, we discuss in detail all the exploration modes.

\subsubsection{Random exploration mode (REM)}
\begin{figure}[htb!]
    \centering
    \includegraphics[width=0.5\textwidth]{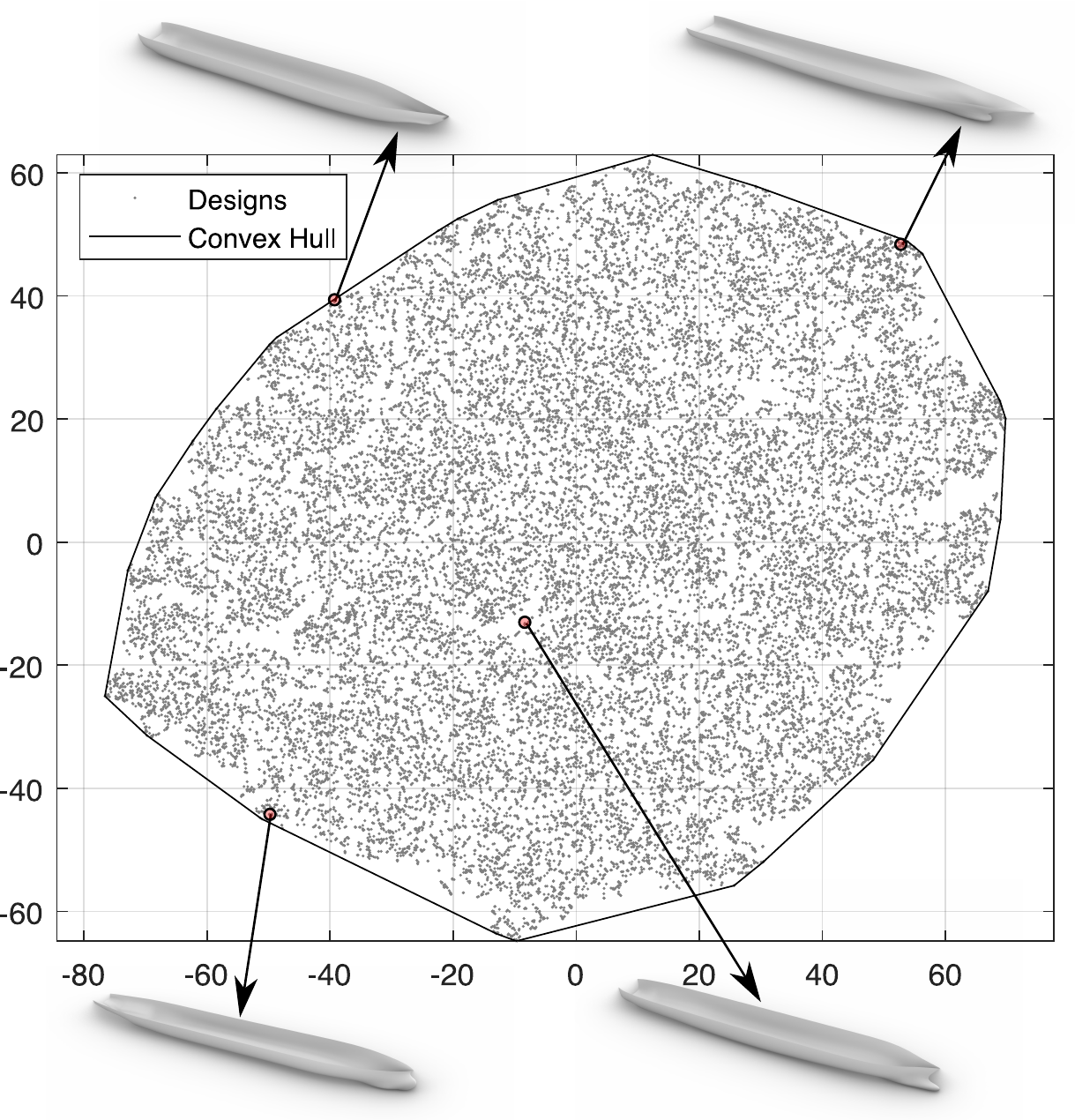}
    \caption{2D t-SEN plot of designs generated from the ShipHullGAN model.}
    \label{tsen}
    \end{figure}

REM is based on a typical random design exploration approach \cite{bole2011interactive,fuchkina2018design,krish2011practical}, where the user manually explores GDS for novel and better-performing designs based on their intuition. However, as GDS has 20 dimensions, the exploration needs to be organised and user-friendly since exploring each of the 20 parameters individually can be cognitively taxing. Therefore, to streamline the exploration process, we first randomly sample a set of 30,000 designs from the GDS that satisfy all the design constraints in Eq. \eqref{shipganOptEq1}. As designers can explore designs well when the dimensionality of the space is low, the sampled designs are projected onto a 2-dimensional space using t-distributed stochastic neighbour embedding (t-SNE) \cite{van2008visualizing}. This statistical method allows for visualising high-dimensional data by giving each data point a location in a 2- or 3-dimensional map that indicates the distribution of designs. The projection of the randomly sampled designs onto a 2-dimensional space is shown in Figure \ref{tsen}, where their boundary is evaluated using the convex hull, shown using a black curve.

During the design exploration process, users can evaluate the $C_w$ value of each design to balance performance and novelty. However, to avoid biasing users towards physics-based designs only, we do not display the performance in real-time. Once a user discovers a novel design, they can evaluate its performance by clicking on the "evaluate $C_w$" button. Each user is randomly assigned a set of 30,000 designs and asked to select 5 preferred designs during the exploration process. The design selection process aims to identify a design that is both novel and optimised. A design may be considered novel if it visually differs from the designs that the user has previously seen, designed, or worked with. An optimised design has the least $C_w$. Therefore, the objective is to find a design that is both novel and optimised, with distinct features and minimal $C_w$.

The design preview window also allows participants to visualise the design in 3D. Users can rotate, zoom, and pan designs to analyse their features thoroughly. Additionally, participants can overwrite a previously selected design. On each design selection, the user is asked what dictated their selection - the form (i.e., design novelty), performance, or a combination of both. Once users select all five designs, they can terminate and conclude this phase of the study.    \begin{figure*}[htbp!]
    \centering
    \includegraphics[width=1\textwidth]{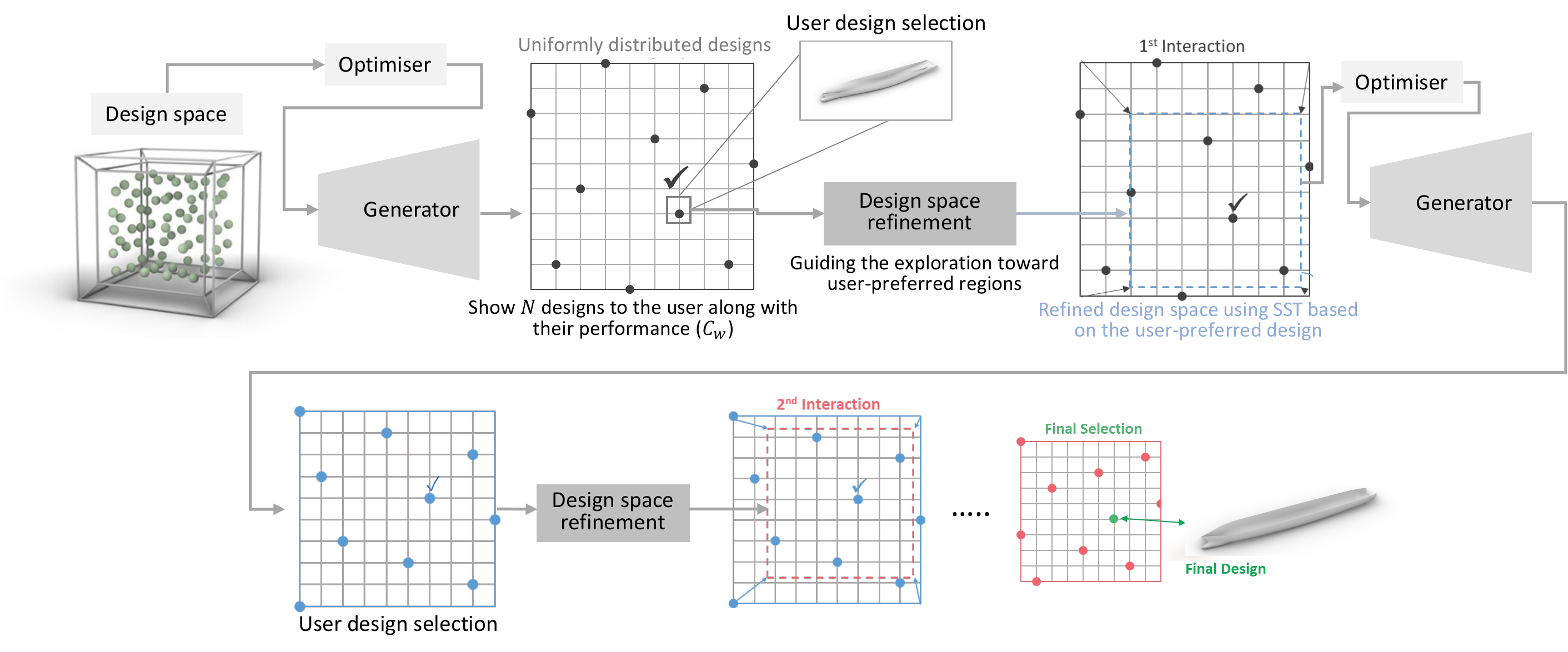}
    \caption{Workflow of semi-automated exploration mode.}
    \label{f1.1}
    \end{figure*}

\subsubsection{Semi-automated exploration mode (SAEM)}
During this mode of exploration, the user and optimiser collaborate to explore GDS for the generation of novel and optimised designs based on the user's intuition and performance. The overall workflow of SAEM is shown in Figure \ref{f1.1}. The optimiser works on exploring a diverse set of optimised designs, while the user induces their preferences to guide the exploration towards user-centred regions of GDS.

Optimisation in this mode is performed based on Khan et al.'s \cite{khan2019genyacht} approach, which provides an innovative way to explore GDS and generate well-diverse design alternatives. This approach commences the design exploration with $N$ number of uniformly distributed designs, where each design represents a particular location in GDS. These designs are then shown to the user along with their $C_w$ values. Afterwards, users select the designs according to the designs' overall form appearance and physics. 

This interaction step allows users to compare designs and make appropriate design decisions. Once the desired hull form is selected, the design space is refined based on the selected design. The refined design space is then imported into the optimiser to generate new $N$ uniformly distributed designs for the next interaction step.

During this process, the designs generated for each interaction should reflect the user’s design selection at the previous interaction, so at the end of the interactive process, the user is able to generate a preferred design. In this work, this is achieved by refining the input design space at each interaction while taking into account the user’s design preference. A Space Shrinking Technique (SST) \cite{khan2019genyacht} is utilised, which detects non-potential regions based on the selected designs and then removes these regions to create a new design space. In other words, at each interaction, SST shrinks the design space towards user-preferred designs and removes regions containing non-preferred designs. This helps the search process to focus the computational effort on the exploration of user-preferred regions of design space. 

The interactive process continues until the user arrives at a design with the desired characteristics. At each design selection, the system asks the user what factors influenced their selection, whether it was performance, form, or a combination of both. The user is permitted to perform between 16-25 interactions, which has been found to be an appropriate number to achieve convergence, meaning that no distinct designs are being created.

\subsubsection{Automated exploration mode (AEM)}

    \begin{figure*}[htb!]
    \centering
    \includegraphics[width=1\textwidth]{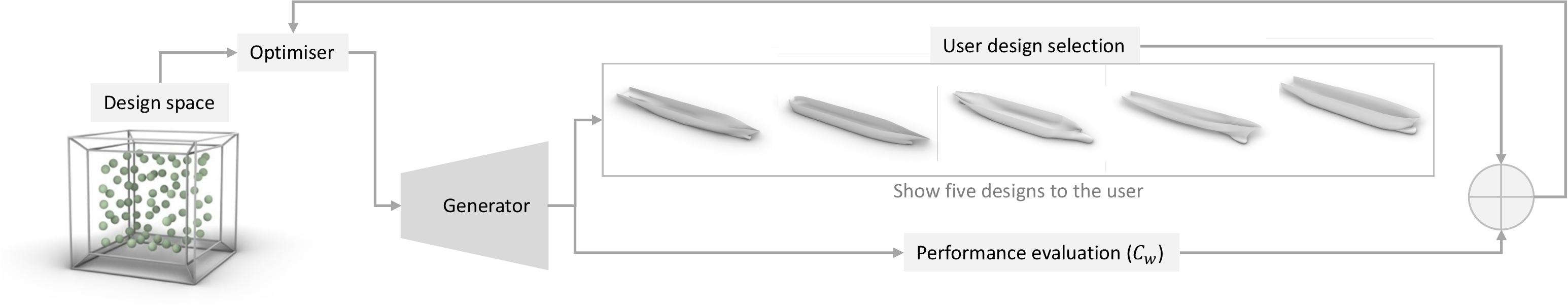}
    \caption{Workflow of automated exploration mode.}
    \label{f2}
    \end{figure*}

This mode of exploration is based on typical shape optimisation \cite{intra_r56}. Its pipeline is shown in Figure \ref{f2}, which connects the GDS, generator (i.e., parametric modeller), and surrogate model for $C_w$ to an appropriate optimiser. During the exploration, the optimiser explores the GDS based on the outcome from the surrogate model, thereby guiding the exploration towards potentially obtaining the global optima while satisfying a given set of constraints.

For the optimisation, we utilised a metaheuristic optimiser, Jay Algorithm (JA), a simple yet efficient approach that does not require any tuning parameters to reach a potentially global solution. JA commences the optimisation with a set of randomly sampled solutions, whose location is improved over a set of iterations. In each iteration, these solutions are moved towards global optima while minimising the following objective function.

\begin{equation}\label{eq_AEM}
\min_{\mathbf{x}\in\mathcal{X}}F=\gamma_1C_w+\gamma_2\sum_{i=1}^n||\mathbf{x}_u-\mathbf{x}_i||
\end{equation}.

The above equation is defined as the weighted sum of two terms. The first term is $C_w$ and the second term is added to induce a notion of user preference during design exploration, which, therefore, is defined as the closeness of the new designs with the previously selected design by the user, $\mathbf{x}_u$. The weights $\gamma_1$ and $\gamma_2$ can be varied between 0 and 1 and set by the user in real-time during exploration. However, initially, we commence the exploration with $\gamma_1=0.7$ and $\gamma_2=0.3$, giving 70\% weightage to $C_w$ and 30\% to the closeness/similarity of newly created designs to the previously selected design.

In our case, since our solver relies on a surrogate model, running many design iterations is not computationally expensive. Therefore, we begin the optimizer with 50 design solutions, which increases the likelihood of finding a good solution. In each iteration, we present the user with the top $n=5$ designs that minimise the objective function in Equation \eqref{eq_AEM}. It is important to note that during the first iteration  $\gamma_2=0$ as there is no preferred design selected by the user. However, starting from the second interaction, participants select a design based on its novelty and performance and adjust the weightage of the objective function accordingly. This process continues in a similar fashion to the previous mode for 16-25 interactions.

\subsection{Population and recruitment}\label{Poprecru}
\begin{figure*}[htb!]
    \centering
    \includegraphics[width=01\textwidth]{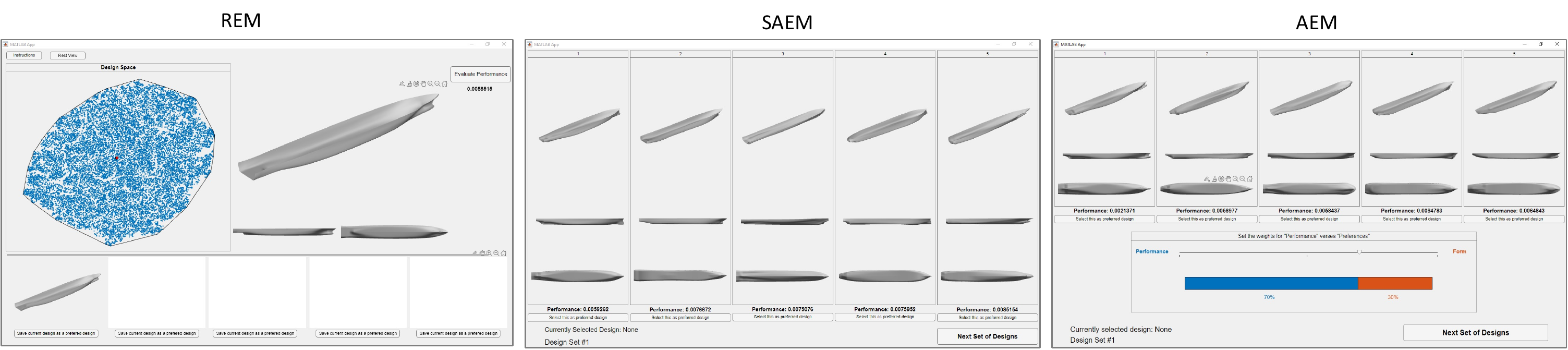}
    \caption{Graphical user interfaces of all three exploration modes.}
    \label{GUI_expMode}
    \end{figure*}

 Figure \ref{GUI_expMode} show the graphical user interface created in MATLAB®\footnote{https://www.mathworks.com} using the above-described exploration approaches. 
In total, 20 participants were recruited for the experiment following a protocol approved by the Institutional Review Board of the University of California and the ethical panel of the University of Strathclyde. All participants were final-year undergraduate students who had taken a Naval architecture course, and on average, they reported 3-4 years of experience in ship design. Participants were offered £30 as compensation for their participation. The average age of the participants was 25, with 30\% female and 70\% male participants. On average, designers reported that they equally value form (i.e. design novelty) and performance in their ship design practice.

The experiments were conducted virtually on Amazon Web Services. Prior to participation, informed consent was obtained from all participants via Google Forms. Participants received an email with step-by-step instructions on the experiment and were assigned 40 minutes to complete it, although they were allowed to take longer. Participants were also informed that there were no right or wrong answers and that their task was to explore the design in a Human-AI design setting using their experience, intuition, and the given directions.

%During the study, data about the design exploration was collected, including all designs considered, the time they were considered, and which designs were designated as favourites. 
The study did not capture any identifying information about the participants. All three modes of exploration were randomly assigned to the participants, meaning that one participant may perform REM first while another participant performs SAEM first. Once the participants completed all three modes of exploration, they were asked to fill out a post-experiment questionnaire designed to gain more insight into the results of the study. The questionnaire contained five questions, which were as follows:

 \begin{enumerate}
     \item[Q1] What mode of exploration helped to:
     \begin{enumerate}
         \item[Q1.1] explore diverse designs
         \item[Q1.2] explore better-performing designs
         \item[Q1.3] explore a mix of diverse and better-performing designs
     \end{enumerate}
     \item[Q2] During the exploration preferred design selection is driven by:
     \begin{enumerate}
         \item[Q2.1] design novelty (i.e., distinctive form features)
         \item[Q2.2] design performance 
         \item[Q2.3] design novelty and performance
     \end{enumerate}
     \item[Q3] The most engaging mode of exploration 
     \item[Q4] The least engaging mode of exploration 
     \item[Q5] Overall preferred mode of exploration
 \end{enumerate}

\section{Results}\label{Results}

In this section, we extensively analyse the outcomes of the user study.

\subsection{Design Histories}
By combining both design histories and final outcomes, it is possible to evaluate the effectiveness of an exploration mode, as described in \cite{girotra2010idea}. In this study, a design history includes:

\begin{enumerate}
\item Overall time spent by participants in each mode.
\item Time spent on each design.
\item Number of designs explored in each mode.
\item Performance of all the explored and user-preferred designs.
\item Indicators for selecting preferred designs: performance, novelty, or a combination of both.
\end{enumerate}

The data relating to these design histories were collected in real-time as the participants performed the study. At the end of the study, the results were automatically sent to the cloud. Among the design histories mentioned above, the key parameters to understand the significance of each mode of exploration are the overall time spent during each mode of exploration, the diversity of the selected designs, and their performance. For example, the most efficient mode of exploration is one in which participants extensively explore the GDS to find diverse yet optimised solutions within a short amount of time. In addition to the above histories, we also measure some mode-specific histories such as the location of designs explored during REM to identify if participants tend to cover the entire design space during exploration. Furthermore, we store the weightage of the two terms of the objective function in Eq. \eqref{eq_AEM} during AEM. These design histories can reveal specific behaviours demonstrated by participants during each mode of exploration \cite{brown2020design}.

\subsection{Analyses of design histories}
Here we first analyse the three key elements of design histories related to the overall time spent, diversity of the explored designs and their quality (i.e., their performance) to gain insight into the behaviour of the participants during the three modes of exploration. 

\subsubsection{Overall time taken}

    \begin{figure}[htb!]
    \centering
    \includegraphics[width=0.45\textwidth]{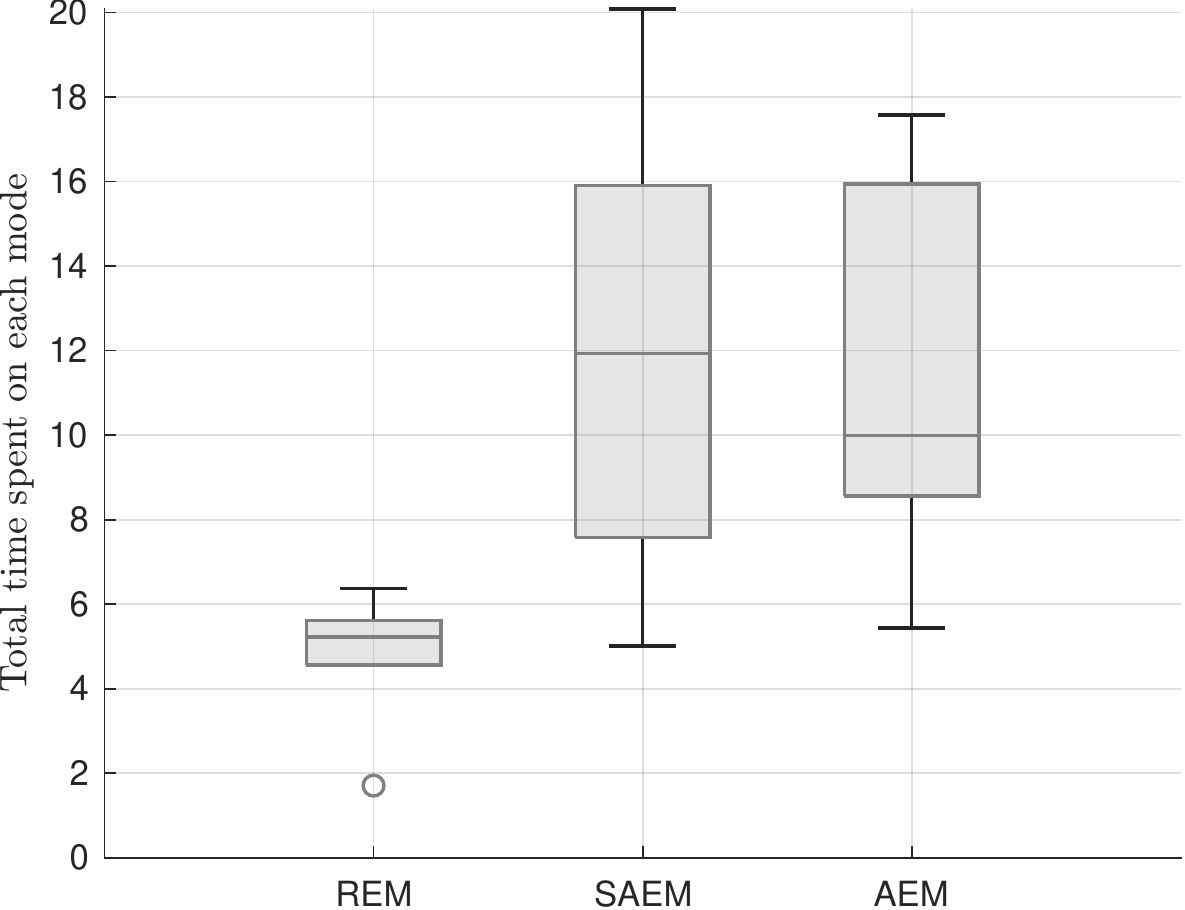}
    \caption{Distribution of total time spent during each mode of exploration.}
    \label{f3}
    \end{figure}

\begin{figure}[htb!]
    \centering
    \includegraphics[width=0.45\textwidth]{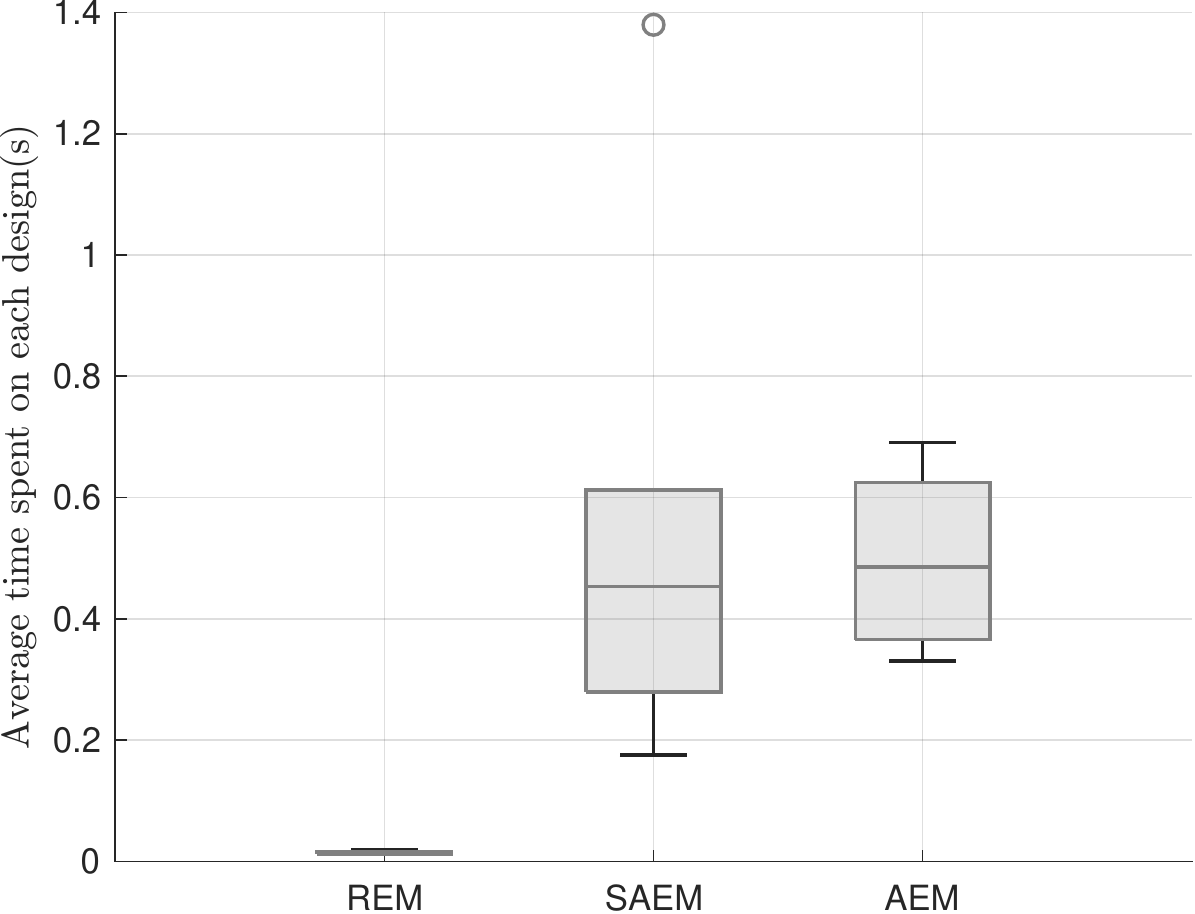}
    \caption{Distribution of average time spent during each design during the REM and average time spent on a set of five designs during each interaction of SAEM and AEM.}
    \label{f4}
    \end{figure}

Figure \ref{f3} shows the total time spent by the participants during each mode of exploration. Interestingly, among the three modes of exploration, participants spent less time in REM, while there was no significant difference between SAEM and AEM. On average, participants completed REM, SAEM, and AEM in 5, 11, and 10 minutes, respectively. It was expected that during REM, participants would take more time to find an innovative and optimised design. However, within REM, participants on average explored 1630 designs within the least amount of time. Another interesting finding was that participants who took less time to complete REM explored more designs, while participants who took more time explored fewer designs. For example, one participant examined 300 designs in 6.4 minutes, whereas another participant explored 6,776 designs in 4.5 minutes. It is important to note that the latter participant's performance can be considered an outlier. Nonetheless, the average time spent on each design, which is shown in Figure \ref{f4}, did reveal an interesting trend: participants who took more time exploring fewer designs spent, on average, more time on each design. The time spent on each design was measured as the time taken to move to a different design from the design that was currently on the viewing window. In other words, it was the time taken when a design was created to the time when it was replaced by a new design. If no new design was created, it meant that the designer was currently analysing the current design, i.e., they were evaluating its performance and/or analysing its feature for novelty. On average, participants spent 1.4 seconds on each design during REM.
    
     \begin{figure}[htb!]
    \centering
    \includegraphics[width=0.45\textwidth]{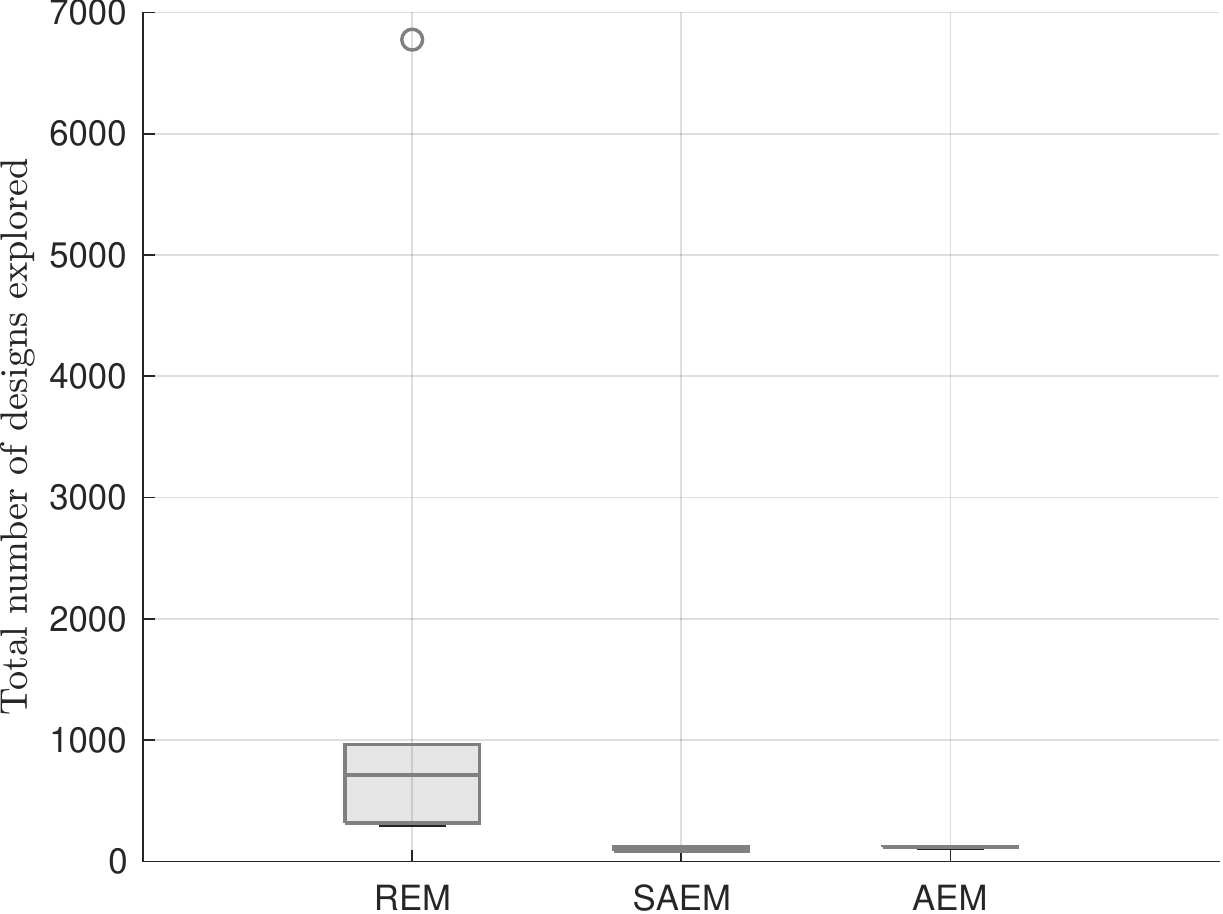}
    \caption{Distribution of a total number of designs explored during each mode of interaction.}
    \label{f5}
    \end{figure}
    
Figure \ref{f5} provides the distribution of the total number of designs explored during each mode of exploration. During REM, participants explored an average of 1630 designs. However, among 20 participants, one participant explored 6,776 designs, which is significantly higher compared to the other participants and can be considered an outlier. If we exclude this outlier, the average number of designs explored by the remaining participants in REM is 601. In contrast, the total number of designs explored during SAEM and AEM was less than REM because, in these modes, participants can explore a set of five designs over 16 to 25 interactions, resulting in a total of 80 to 125 designs. As explained earlier, this number was chosen based on pre-analysis to ensure that participants could explore the design space without experiencing cognitive overload. On average, participants explored 106 and 118 designs in SAEM and AEM, respectively. These results indicate that within SAEM, participants were able to quickly scan the GDS with the least number of designs, taking approximately the same time as in AEM, which has the highest level of design exploration automation.

 \subsubsection{ Diversity of preferred designs}
In this subsection, we analyse the diversity of the preferred designs during the three modes of exploration. For REM, the diversity is measured between the five final selected designs, and for SAEM and AEM, it is measured between the design selected as preferred designs during each interaction. This analysis addresses the effect of partially performance-driven design exploration and its impact on creativity for the creation of novel hull forms. While creativity and design novelty can be defined in different ways, it is generally assumed that measurements of diversity correspond with increased relative freedom, while the tendency towards standard solutions indicates less creative freedom \cite{brown2020design}. The diversity measure aims to specifically understand if giving the performance as a criterion for exploration biases participants to increase novelty or if it influences participants to still focus only on the performance, as in the typical design exploration setting.

Diversity in this work is evaluated with the sparseness at the centre (SC)~\cite{brown2019quantifying} criterion, which measures the average distance of the centroidal design, $\mathbf{x}^{centroid}$, of the preferred design, to the preferred designs resulting during the exploration of GDS. 

\begin{equation}\label{shipgan_eq_D}
    SC  = \frac{1}{n}\sum_{i=1}^n{||\mathbf{x}_{centroid} - \mathbf{x}_i||}_2
\end{equation}

\begin{figure}[htb!]
    \centering
    \includegraphics[width=0.45\textwidth]{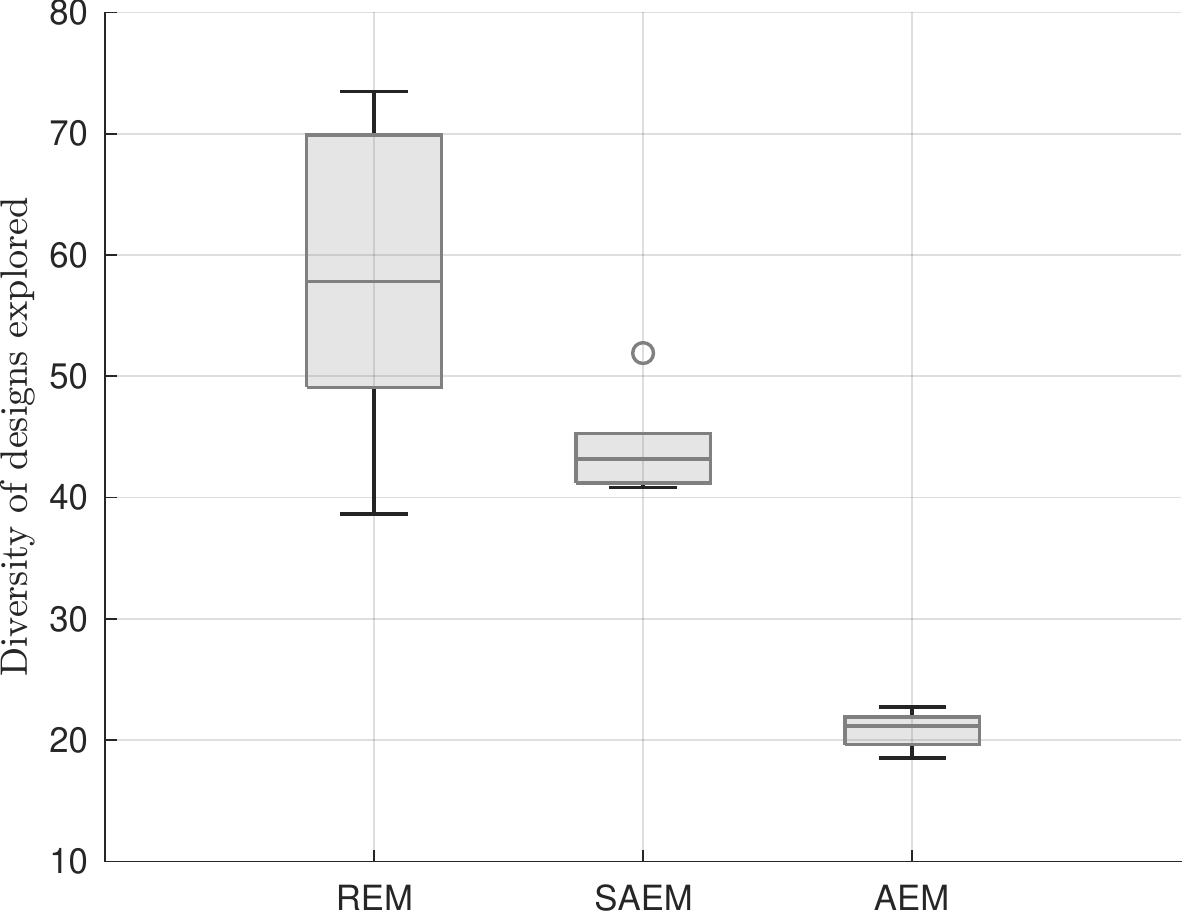}
    \caption{Distribution of the diversity of designs explored by participants in all three modes of exploration.}
    \label{f6}
    \end{figure}

Although the absolute units of SC measurement are meaningless, as they represent the distance between designs, the relative values from the different modes of exploration provide a worthwhile comparison. Figure \ref{f6} shows the SC measure of the designs explored by the participants in all three modes of exploration. It is interesting to note that the diversity of the preferred designs in REM is significantly higher compared to the other two modes. AEM has significantly lower diversity, indicating that designs are highly influenced by performance without much focus on diversity, even when the objective function includes a term to induce a human preference for novelty (see Eq. \eqref{eq_AEM}).

\subsubsection{Performance of selected designs}
     \begin{figure}[htb!]
    \centering
    \includegraphics[width=0.45\textwidth]{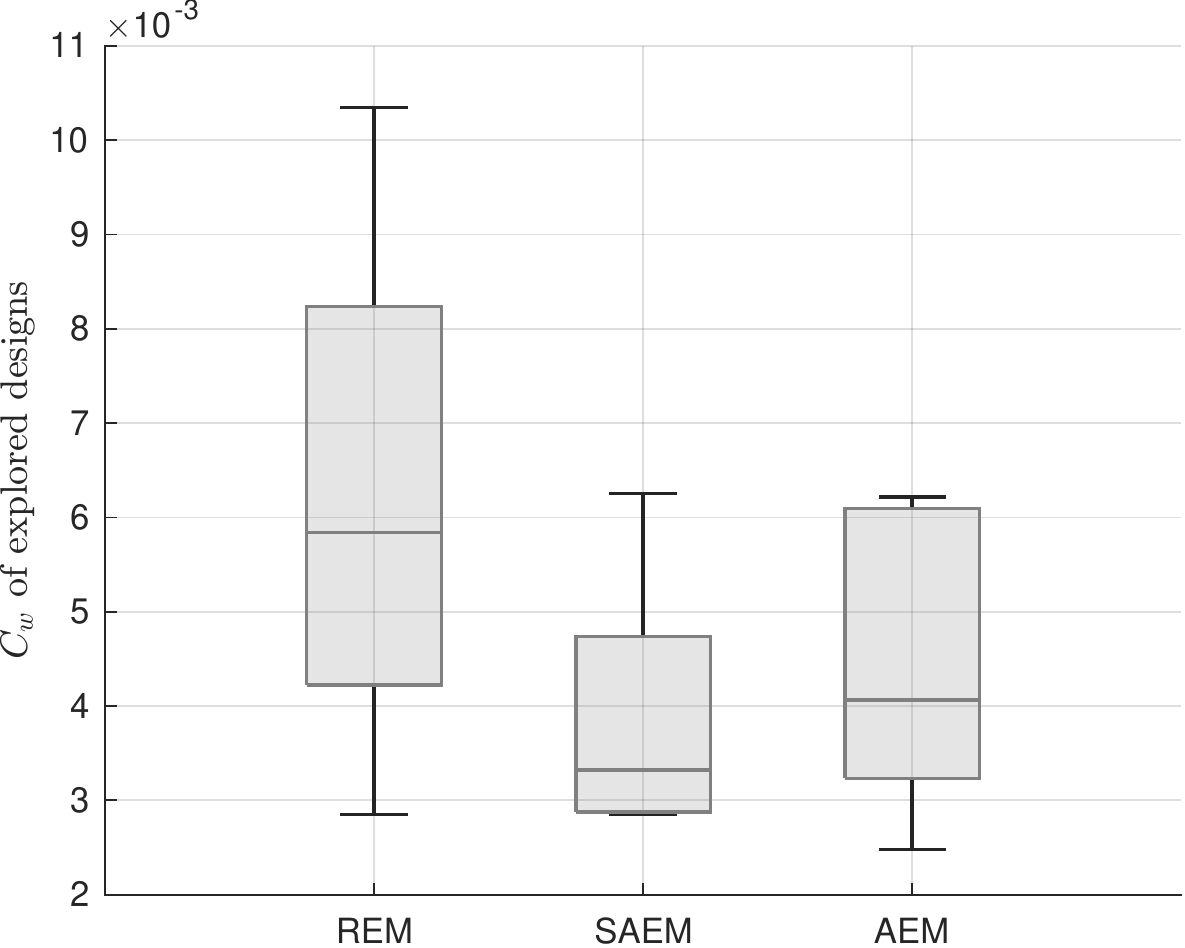}
    \caption{Distribution of $C_w$ values of the design explored during the three modes of explorations.}
    \label{f7}
    \end{figure}
Figure \ref{f7} shows the average values of $Cw$ of the preferred design resulting from all three modes. It is noteworthy that designs resulting from SAEM, on average, perform better compared to AEM, which is highly performance-driven. However, designs resulting from REM are diverse but do not perform well. In conclusion, participants find better performing and diverse designs with SAEM while exploring fewer designs compared to AEM and REM.

\subsubsection{Performance vs novelty}
During the three modes of exploration, most participants tended to select the preferred design based on both performance and form novelty. However, in REM, participants cared more about form novelty, while in the other two modes, they prioritised performance. Interestingly, the indication towards performance was higher in SAEM, which could be the reason for the better-performing preferred designs resulting from SAEM.

Another point worth noting is that at the start of the study, we asked the participants to give their opinion on whether they care more about performance or novelty during a typical design process. On average, they indicated an equal preference for both novelty and performance.

\subsection{Survey results}

     \begin{figure}[htb!]
    \centering
    \includegraphics[width=0.45\textwidth]{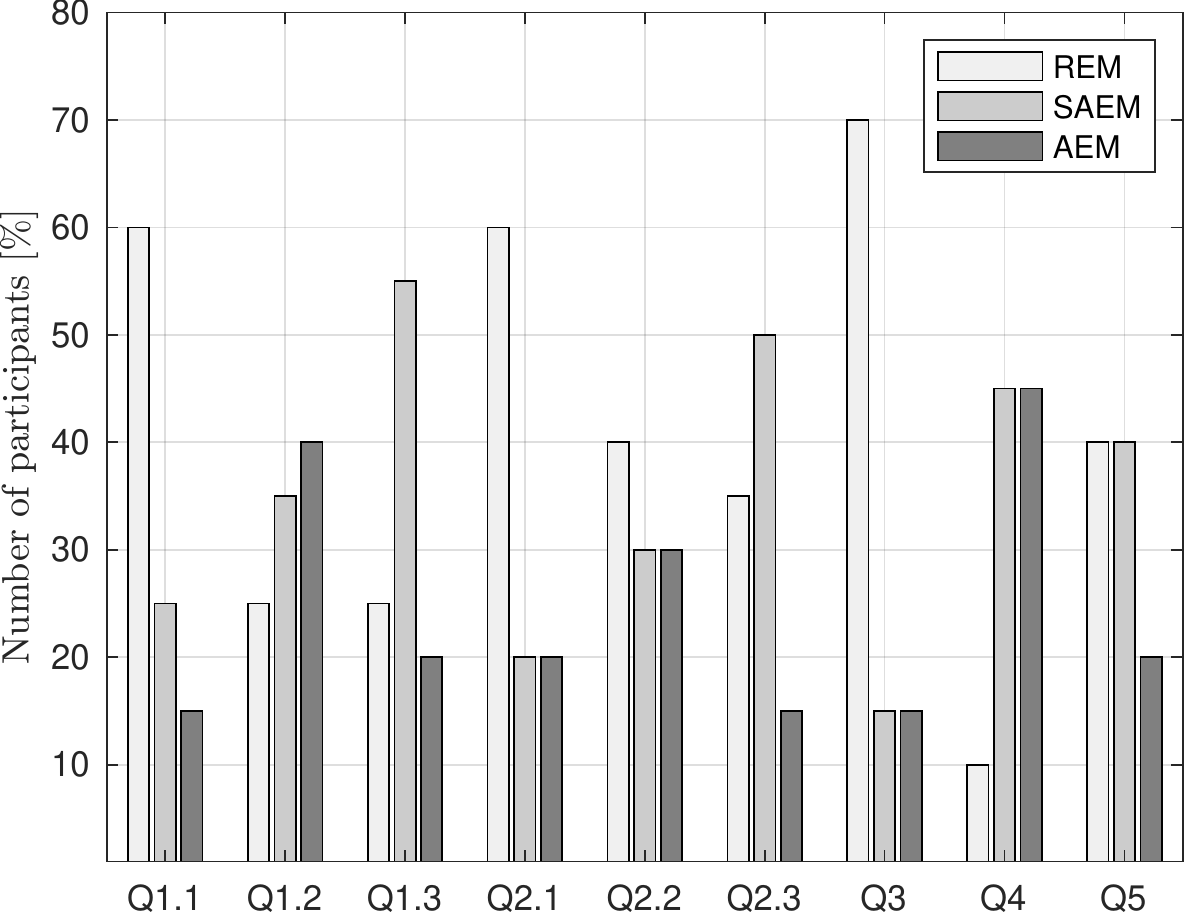}
    \caption{Summary of the questionnaire results from all the participants.}
    \label{f8}
    \end{figure}
In this subsection, we discuss the results of the questionnaire conducted to evaluate participants' perceptions of RME, SAME and AME exploration modes discussed in Section \ref{Poprecru}. The results of this questionnaire are shown in Figure \ref{f8}. It can be seen that in Q1.1, 60\% of participants reported finding the most novel design ideas within REM, while only 15\% found AEM to be useful for discovering novel designs. However, in Q1.2, 40\% of the participants believed that AEM produced better-performing designs. The remaining 35\% and 25\% of participants found REM and SAEM, respectively, to be more effective at producing better-performing designs. In response to Q1.3, which asked about the mode that provided the exploration of both diverse and better-performing designs, 55\% of participants preferred SAEM, while 25\% and 20\% preferred REM and AEM, respectively.

From Figure \ref{f8} it can be seen that in Q2.1, 60\% of participants indicated that novelty was the driver of design selection in REM. Additionally, in Q2.2, 40\% of participants indicated that performance was also a driver of design selection in REM, although this result did not deviate significantly from those of SAEM and AEM. The results of Q2.3 were particularly interesting, as 50\% of participants found SAEM to be a mode where design selection was driven equally by both performance and novelty. It is worth noting that the trend observed in the Q2 questions is consistent with that of the Q1 questions. 

For Q3, Q4 and Q5, the results in Figure \ref{f8} show that 70\% of the participants found REM to be the most engaging mode, whereas only 10\% of the participants indicated it as the least engaging mode. Perhaps these are the participants who value design performance significantly more than design novelty. Overall, participants least favoured the AEM, mainly due to the lack of design novelty.

\section{Discussion}
This section provides a detailed discussion of the key findings of this study, specifically focused on identifying an effective exploration approach for GDS.
\subsection{To what extent is each exploration mode effective in achieving diverse, novel and better-performing designs?}
From the results discussed in Section \ref{Results}, it can be concluded that REM provides a user-engaging approach that results in the exploration of novel design solutions within the shortest possible design exploration time. Interestingly, even with a short exploration time, participants are able to scan the GDS effectively and find a diverse set of design alternatives. Although designs resulting from this mode are diverse, as expected, they are not efficient from a performance perspective. On the other hand, AEM, which is primarily driven by performance, has the least diversity. One would expect that even if the designs are not diverse, they must perform better as the optimisation is solely driven by performance. However, designs resulting from AEM have slightly lower performance on average compared to the designs resulting from SAEM. 

In conclusion, while design spaces resulting from typical parametric approaches may yield better-performing designs, the key benefit of GDSs lies not only in better performance but also in the ability to generate novel designs that possess non-conventional features and do not currently exist in the market. To fully leverage the potential of GDSs, it is necessary to focus not only on enriching these spaces but also on exploring them in an effective manner. This study highlights that the commonly used random exploration (REM) and optimisation-based exploration (AEM) approaches are not optimal for GDSs. REM prioritises novelty, while AEM prioritises performance. To strike a balance between these objectives, hybrid and intuitive exploration approaches such as SAEM are needed. SAEM involves both the user and the optimiser at the same level, where users leverage their design expertise to explore novel solutions, and the optimiser focuses on enhancing the performance of the user-preferred designs.

\subsection{Which factor is the key consideration for each exploration
mode: form or performance?}
Moreover, this study revealed that although participants initially stated they aim to balance both novelty and performance in their design tasks. However, during the design exploration in REM, SAEM and AEM they tended to prioritise performance over novelty. This may be due to various factors. If this trend persists, users may be biased toward prioritising performance and fail to utilise GDSs to their full potential. However, this fact is also centred on the type of exploration approach used. For example, in REM, design exploration and selection of preferred design was driven by the form, whereas in AEM design selection was mainly driven by the performance. However, SAEM, aimed at balancing both performance and novelty, does well to engage participants in design exploration. Therefore, designs resulting from this approach are diverse as well as better performing. Furthermore, in addition to hybrid exploration methods, there is a need for more engaging design interfaces. Our questionnaire results indicate that participants found REM to be more engaging compared to other exploration modes.

In summary, from this study, it can be concluded that as the design space are become more and more diverse, thanks to generative models, we need also an innovative approach for their efficient exploration, as the traditional way of design exploration, made originally for the narrow design spaces, cannot be beneficial to truly exploit the potential of GDSs.

\section{Concluding remarks}
In this work, we aimed to evaluate the effectiveness of different design exploration approaches for exploring generative design spaces resulting from generative models such as generative adversarial networks. To achieve this, we constructed a generative design space for the ship hull design and optimisation problem. We trained a custom generative model on a large dataset of physically and geometrically validated designs and then used the generator component of the model as a parametric modeller to generate a diverse 20-dimensional design space. We explored this space using three different approaches: REM, SAEM, and AEM, each with different levels of user involvement and algorithmic autonomy.

The REM is a random exploration mode in which the user explores a 2-dimensional projection of the generative design space. SAEM is a mode that is spontaneously driven by both the user and the optimiser with the same level of involvement. In this mode, the optimiser focuses on exploring a diverse set of uniformly distributed and optimised designs (i.e., designs with low $C_w$) from the generative design space, while the user directs the exploration towards the region of design space containing their preferred designs. AEM is a typical shape optimisation mode in which the design space is connected to an optimiser and performance evaluation code that guides the optimiser in finding a global optimum. To incorporate a user's preference, the objective function for this mode is the weighted sum of $C_w$ and the similarity of newly created designs to the user's previously selected designs. 

The results of this study showed that the highest design diversity occurred during REM, followed by SAEM and AEM, whereas better-performing designs were found within AEM and SAEM. However, SAEM outperforms REM and AEM in terms of exploring designs that have a significantly high trade-off between novelty and performance. The study results also showed that participants are adept at exploring novelty and that their subconscious directs them to prefer novel design alternatives. However, when performance is brought into the exploration, they immediately tend to select based on performance.

\subsection{Future work}
In the future, we aim to scale up our investigation by enriching the population of the subjects involved with a) designers covering the whole spectrum of expertise (low to high), b) design-users covering the whole lifecycle, e.g. shipyards, ship-owners, operators, and c) designers acting in other transportation (automotive, aerospace) industries.

\section*{Acknowledgements}
 This work received funding from:
\begin{enumerate}
\item the Royal Society under the HINGE (Human InteractioN supported Generative modEls for creative designs) project via their International Exchanges 2021 Round 2 funding call, and
\item the European Union’s Horizon 2020 research and innovation programme under the Marie Skłodowska-Curie grant GRAPES (learninG, pRocessing And oPtimising shapES) agreement No 860843.
 \end{enumerate}
\bibliographystyle{elsarticle-num}
 \bibliography{refs}

\begin{thebibliography}{10}
\expandafter\ifx\csname url\endcsname\relax
  \def\url#1{\texttt{#1}}\fi
\expandafter\ifx\csname urlprefix\endcsname\relax\def\urlprefix{URL }\fi
\expandafter\ifx\csname href\endcsname\relax
  \def\href#1#2{#2} \def\path#1{#1}\fi

\bibitem{intra_r56}
K.~Kostas, A.~Ginnis, C.~Politis, P.~Kaklis, Ship-hull shape optimization with
  a {T}-spline based {BEM}--isogeometric solver, Computer Methods in Applied
  Mechanics and Engineering 284 (2015) 611--622.

\bibitem{chen2021padgan}
W.~Chen, F.~Ahmed, Padgan: Learning to generate high-quality novel designs,
  Journal of Mechanical Design 143~(3) (2021).

\bibitem{regenwetter2022deep}
L.~Regenwetter, A.~H. Nobari, F.~Ahmed, Deep generative models in engineering
  design: A review, Journal of Mechanical Design 144~(7) (2022) 071704.

\bibitem{yang2020physics}
L.~Yang, D.~Zhang, G.~E. Karniadakis, Physics-informed generative adversarial
  networks for stochastic differential equations, SIAM Journal on Scientific
  Computing 42~(1) (2020) A292--A317.

\bibitem{chaudhari2023evaluating}
A.~M. Chaudhari, D.~Selva, Evaluating designer learning and performance in
  interactive deep generative design, Journal of Mechanical Design 145~(5)
  (2023) 051403.

\bibitem{khan2017novel}
S.~Khan, E.~Gunpinar, K.~M. Dogan, A novel design framework for generation and
  parametric modification of yacht hull surfaces, Ocean Engineering 136 (2017)
  243--259.

\bibitem{ShipHullGAN_2023}
S.~Khan, K.~Goucher-Lambert, K.~Kostas, P.~Kaklis, Ship{H}ull{GAN}: {A} generic
  parametric modeller for ship hull design using deep convolutional generative
  model, Computer Methods in Applied Mechanics and Engineering 411 (2023)
  116051.

\bibitem{charisi2022early}
N.~D. Charisi, H.~Hopman, A.~Kana, Early-stage design of novel vessels: How can
  we take a step forward?, in: SNAME 14th International Marine Design
  Conference, OnePetro, 2022.

\bibitem{nowacki2010five}
H.~Nowacki, Five decades of computer-aided ship design, Computer-Aided Design
  42~(11) (2010) 956--969.

\bibitem{gaspar2012addressing}
H.~M. Gaspar, D.~H. Rhodes, A.~M. Ross, S.~Ove~Erikstad, Addressing complexity
  aspects in conceptual ship design: a systems engineering approach, Journal of
  Ship Production and Design 28~(04) (2012) 145--159.

\bibitem{ebrahimi2021influence}
A.~Ebrahimi, P.~O. Brett, S.~O. Erikstad, B.~E. Asbj{\o}rnslett, Influence of
  ship design complexity on ship design competitiveness, Journal of Ship
  Production and Design 37~(03) (2021) 181--195.

\bibitem{papanikolaou2010holistic}
A.~Papanikolaou, Holistic ship design optimization, Computer-Aided Design
  42~(11) (2010) 1028--1044.

\bibitem{khan2017customer}
S.~Khan, E.~Gunpinar, M.~Moriguchi, Customer-centered design sampling for cad
  products using spatial simulated annealing, Proceedings of CAD 17 (2017)
  100--103.

\bibitem{GinnisEtAl2011}
A.~Ginnis, K.~Kostas, C.~Feurer, K.~Belibassakis, T.~Gerostathis, C.~Politis,
  P.~Kaklis, A {CATIA}\textregistered ship-parametric model for isogeometric
  hull optimization with respect to wave resistance, in: Proceedings of ICCAS
  2011 conference, Trieste 20-22 September, Italy, 2011.

\bibitem{khan2022modiyacht}
S.~Khan, E.~Gunpinar, K.~Mert~Dogan, B.~Sener, P.~Kaklis, Modi{Y}acht:
  Intelligent cad tool for parametric, generative, attributive and interactive
  modelling of yacht hull forms, in: SNAME 14th International Marine Design
  Conference, OnePetro, 2022.

\bibitem{khan2019genyacht}
S.~Khan, E.~Gunpinar, B.~Sener, Genyacht: An interactive generative design
  system for computer-aided yacht hull design, Ocean Engineering 191 (2019)
  106462.

\bibitem{DKaklisEtAl2023SOME}
D.~Kaklis, T.~Varelas, I.~Varlamis, P.~Eirinakis, G.~Giannakopoulos, C.~V.
  Spyropoulos, From steam to machine: Emissions control in the shipping 4.0
  era, in: The 8th International Symposium on Ship Operations, Management \&
  Economics (SOME), Society of Naval Architects and Marine Engineers (SNAME),
  2023, pp. 1--12.

\bibitem{citaristi2022united}
I.~Citaristi, United nations conference on trade and, in: The Europa Directory
  of International Organizations 2022, Routledge, 2022, pp. 177--181.

\bibitem{joung2020imo}
T.-H. Joung, S.-G. Kang, J.-K. Lee, J.~Ahn, The imo initial strategy for
  reducing greenhouse gas (ghg) emissions, and its follow-up actions towards
  2050, Journal of International Maritime Safety, Environmental Affairs, and
  Shipping 4~(1) (2020) 1--7.

\bibitem{khan2022shape}
S.~Khan, P.~Kaklis, A.~Serani, M.~Diez, K.~Kostas, Shape-supervised dimension
  reduction: Extracting geometry and physics associated features with geometric
  moments, Computer-Aided Design 150 (2022) 103327.

\bibitem{khan2022geometric}
S.~Khan, P.~Kaklis, A.~Serani, M.~Diez, Geometric moment-dependent global
  sensitivity analysis without simulation data: application to ship hull form
  optimisation, Computer-Aided Design 151 (2022) 103339.

\bibitem{bertram2011practical}
V.~Bertram, Practical ship hydrodynamics, Elsevier, 2011.

\bibitem{schulz2018tutorial}
E.~Schulz, M.~Speekenbrink, A.~Krause, A tutorial on gaussian process
  regression: Modelling, exploring, and exploiting functions, Journal of
  Mathematical Psychology 85 (2018) 1--16.

\bibitem{khan2021regional}
S.~Khan, P.~Kaklis, From regional sensitivity to intra-sensitivity for
  parametric analysis of free-form shapes: Application to ship design, Advanced
  Engineering Informatics 49 (2021) 101314.

\bibitem{bassanini1994wave}
P.~Bassanini, The wave resistance problem in a boundary integral formulation,
  Surv Math Ind 4 (1994) 151--194.

\bibitem{khan2021physics}
S.~Khan, A.~Serani, M.~Diez, P.~Kaklis, Physics-informed feature-to-feature
  learning for design-space dimensionality reduction in shape optimisation, in:
  AIAA scitech 2021 forum, 2021, p. 1235.

\bibitem{bole2011interactive}
M.~Bole, Interactive hull form transformations using curve network deformation,
  Ship Technology Research 58~(1) (2011) 46--64.

\bibitem{fuchkina2018design}
E.~Fuchkina, S.~Schneider, S.~Bertel, I.~Osintseva, Design space exploration
  framework, in: eCAADe, Vol.~36, 2018, pp. 367--376.

\bibitem{krish2011practical}
S.~Krish, A practical generative design method, Computer-Aided Design 43~(1)
  (2011) 88--100.

\bibitem{van2008visualizing}
L.~Van~der Maaten, G.~Hinton, Visualizing data using t-{SNE}., Journal of
  machine learning research 9~(11) (2008).

\bibitem{girotra2010idea}
K.~Girotra, C.~Terwiesch, K.~T. Ulrich, Idea generation and the quality of the
  best idea, Management science 56~(4) (2010) 591--605.

\bibitem{brown2020design}
N.~C. Brown, Design performance and designer preference in an interactive,
  data-driven conceptual building design scenario, Design studies 68 (2020)
  1--33.

\bibitem{brown2019quantifying}
N.~C. Brown, C.~T. Mueller, Quantifying diversity in parametric design: a
  comparison of possible metrics, AI EDAM 33~(1) (2019) 40--53.

\end{thebibliography}

\end{document}